%% file: main.tex
\documentclass[runningheads]{llncs}
\usepackage{graphicx}
\usepackage{graphics}
\usepackage{wrapfig}
\usepackage{times}
\usepackage{multirow}
\usepackage{subfigure}
\usepackage{float}
\graphicspath{ {figs/} }
\usepackage{siunitx}
\usepackage{mathptmx}
\usepackage{amsmath} 
\usepackage{amssymb}  
\usepackage{bm}
\usepackage{hyperref}
\usepackage{marvosym}

\begin{document}
\title{Hybrid Generative Models for Two-Dimensional Datasets}
\titlerunning{Hybrid Generative Models}
%
\author{Hoda Shajari\textsuperscript{\Letter},
Jaemoon Lee, 
Sanjay Ranka,
Anand Rangarajan}
\authorrunning{H. Shajari et al.}
%
\institute{University of Florida, Gainesville, FL 32611-6120, USA \\
\email{\{shajaris,j.lee1,sranka,anandr\}@ufl.edu}\\
}
\maketitle              
\begin{abstract}
\input{abstract}
\end{abstract}
\input{intro}
\input{previous}

\input{theory}

\input{approach}

\input{experiments}

\input{discussion}
\input{conclusion}

%
%
%
%

\bibliographystyle{splncs04}
\vspace{-0.2 cm}
\bibliography{reference}

\end{document}

%% file: abstract.tex
Two-dimensional array-based datasets are pervasive in a variety of domains. Current approaches for generative modeling have typically been limited to conventional image  datasets and performed in the pixel domain which does not explicitly capture the correlation between pixels. Additionally, these approaches do not extend to scientific and other applications where each element value is continuous and is not limited to a fixed range.
In this paper, we propose a novel approach for generating two-dimensional datasets by moving the computations to the space of representation bases and show its usefulness for two different datasets, one from imaging and another from scientific computing. The proposed approach is general and can be applied to any dataset, representation basis, or generative model. We provide a comprehensive performance comparison of various combinations of generative models and representation basis spaces. We also propose a new evaluation metric which captures the deficiency of generating images in pixel space. 

\keywords
Generative Models, Image Representation Bases, Normalizing Flows, Independent Component Analysis, Generative Adversarial Networks

%% file: intro.tex
\section{Introduction}

The high volume and unique requirements of scientific image datasets necessitate the development of novel approaches for data modeling. The bedrock assumption of all modeling methodologies is the existence of spatiotemporal homogeneities in the data which can be exploited. However, in contrast to two-dimensional image modeling,
scientific data are underpinned by unusual geometries and topologies. This ``exotic setting” has to be leveraged and addressed by machine learning methods in their quest to find homogeneities which in turn can be efficiently exploited using representation bases. 
Additionally, unlike image datasets where pixel values are discrete and within a certain range, the elements of scientific datasets are continuous and can vary for each data point. In this paper, we propose a novel approach for modeling the probability distribution of two-dimensional datasets while developing a new measure for evaluating the models. The proposed approach can be applied to image and scientific datasets, with elements that are either discrete or continuous valued. 
Generative models in machine learning have drawn significant attention with many applications in different fields, including, but not limited to, computer vision, and physics-based simulations for scientific datasets. The importance of generative modeling and approximating data distributions stems from the fact that unlabeled data are relatively abundant compared to labeled data, and this has applications in density estimation, outlier detection, and reinforcement learning. Deep generative modeling also has emerged during the bloom of deep learning and takes advantage of advances in computational power \cite{kang2020generating,kingma2013auto}. However, these models have not leveraged classical methods of data representation. These models usually learn the probability distribution of the images directly in pixel space, which is costly and inefficient while ignoring 50 years of image representation bases used in the compression literature. Furthermore, learning the distribution of the data in pixel space does not leverage the correlation information among pixels. 

Representation basis techniques aim to transform data in such a way that useful aspects of data, for example statistical properties, are captured in the transformed space. Principal Component Analysis (PCA), Independent Component Analysis (ICA), and tensor decompositions using the higher order SVD (which we henceforth encapsulate as the Tucker decomposition for the sake of convenience) are among the widely used methods in this area. The other utility of representation bases is dimensionality reduction, which can be considered as a kind of lossy compression, i.e., it is possible to represent the data with a subset of coefficients with desired accuracy. Therefore, dimensionality reduction also brings a compression aspect into our approach.   

We propose an approach to integrate image representation basis techniques in generative image modeling and perform a comparison among three generative models---generative adversarial networks (GANs), normalizing flows (NFs), and Gaussian mixture modeling (GMM)---and analyze their performance. The results suggest that this is a promising direction to pursue for efficient two-dimensional dataset generative modeling, in particular for applications where resources are scarce and speed of training matters. We summarize the contributions of our work below:
\vspace{-0.3 cm}
\begin{itemize}
    \item We propose an approach for two-dimensional datasets which exploits representation bases to capture the correlation among elements explicitly and therefore makes the generation process fast and efficient.
    This approach is general and can be applied to image and scientific datasets where the underlying data are respectively discrete with a fixed range and continuous with a free range.
    \item We propose a new quantitative metric to compare the performance of our approach for different choices of generative models and representation bases. This metric seems to capture the quality of the learned probability distribution better than conventional metrics, especially for scientific data.
\end{itemize}
\vspace{-0.155 cm}
The rest of the paper is outlined as follows. In section~\ref{sec:previous-work}, we cover previous work on generative models utilizing representation bases or compression concepts. In section~\ref{sec:representation bases}, we provide an overview of representation basis approaches used in this work. In section~\ref{sec:methodology}, our methodology for hybrid generative modeling is outlined. Implementation details and experiments are discussed in section~\ref{sec:experiments}. Section~\ref{sec:conclusion} concludes the paper. 

%% file: previous.tex
\section{Previous Work}
\vspace{-0.2 cm}
\label{sec:previous-work}
Learning the probability distribution of datasets is a long-standing problem, and generative models in machine learning constitute an important class of models with a rich literature. The Gaussian mixture model (GMM) is one of the important and classical models for generative modeling \cite{reynolds2009gaussian} while deep generative models rely on multilayer perceptrons (with deep architectures) for learning the data distribution.

Despite the importance of image representation bases and their abundant application in the compression literature, there are very few approaches which learn image probability distributions by marrying image representation bases (thereby moving away from pixel space) and deep learning-based generative models. Our proposed work therefore bridges the gap between image representation and deep learning-based image generation. As we will demonstrate in section~\ref{sec:methodology}, our approach combines parametric and nonparametric modules where the parametric component is based on representation bases, while the nonparametric module is based on machine learning methods. 

Generative Latent Optimization (GLO) \cite{bojanowski2017optimizing} was proposed as a deep generative model which learns a deep CNN generator to map latent vectors to data by a Laplacian pyramid loss function while forcing latent representations to lie on a unit sphere. In this model, however, it is not possible for the generator to randomly sample from a known distribution. Implicit Maximum Likelihood Estimation (IMLE) \cite{li2018implicit} uses a non-adversarial approach for discovering the mapping between two densities. In this model, latent variables are mapped into image space via a generator and for each training image, the nearest generated image is found such that the $\ell_2$ distance between the image and mapping is minimized and the generator is repeatedly optimized via a nearest neighbor-based loss. IMLE optimization is costly and the generated images are typically blurry. The work in \cite{hoshen2019non} proposed the idea of combining GLO and IMLE to learn a mapping for projecting images into a spherical latent space and learning a network for mapping sampled points from latent space to pixel image space in a non-adversarial fashion. Generative Latent Flow \cite{xiao2019generative} learns the latent space of data via an autoencoder and then maps the distribution of latent variables to i.i.d. noise distributions. In the wavelet domain, SWAGAN \cite{gal2021swagan} proposed a wavelet-based progressive GAN for image generation which improves visual quality by enforcing a frequency-aware latent representation.  Our proposed method is different from the aforementioned approaches because it can be incorporated into any generative model and therefore allows for sampling from a known distribution whenever necessary. Furthermore, it has a parametric module and when used within a GAN architecture, it is trained via an adversarial loss. 

At the intersection of compression and deep generative models, Agustsson et al. \cite{agustsson2019generative} proposed a framework based on GANs for generating images at lower bitrates. The model learns an encoder which includes a quantization module which is trained in combination with a multiscale discriminator. Kang et al. \cite{kang2020generating} proposed a framework for generating JPEG images via GANs. They proposed a generator with different layers for chroma sampling and residual blocks. Our approach is also different from these approaches because image compression concepts are directly used in the form of representation bases (like Tucker etc.) coupled with dimensionality reduction.

The Tucker decomposition \cite{tucker1966some}, PCA \cite{jolliffe2016principal}, and ICA \cite{hyvarinen2000independent} are among the widely-used approaches which linearly transform data into a new space where data can be presented in a more structured way and more efficiently represented. Dimensionality reduction is an important byproduct of representation bases. Choosing a subset of coefficients corresponding to the representation can result in data compression and has been extensively used in the literature. As discussed earlier, our use of these methods is for the purpose of converting from the original data to a new space that captures the correlation between elements as well as providing an efficient representation for further processing. Therein lies the novelty of our work. As far as we know, this is the first work that conducts a comprehensive comparison at the intersection of generative models and representation bases while leveraging recent advances in adversarial learning.

%% file: theory.tex
\section{Representation Bases}
\label{sec:representation bases}
Finding a proper representation of random multivariate data is a key to many domains \cite{hyvarinen2000independent}. Linear transformations have specifically been of interest due to their conceptual and computational simplicity. The following techniques have been used in our work:
\begin{enumerate}
    \item 
{\bf Principal Component Analysis (PCA):}
PCA linearly transforms the data by discovering orthogonal projections of high variance. Given a set of vectors $\{x_{i}\}_{i=1}^{N}, \: x_{i}\in \mathbb{R}^{D}$, a correlation matrix is computed as $C = \Sigma_{i=1}^{N}{x_{i}x_{i}^{T}}$, which has the following eigen decomposition: $C = E\Lambda E^{T}= E\Lambda^{\frac{1}{2}}\Lambda^{\frac{1}{2}T}E^{T} = E\Lambda^{\frac{1}{2}}(E\Lambda^{\frac{1}{2}})^{T}$, where $\Lambda$ is a diagonal matrix of eigenvalues, and their corresponding eigenvectors constitute the columns of $E$. 
Dimensionality reduction is performed by projecting data onto eigenvectors corresponding to the first $d$ maximum eigenvalues, which captures the maximum variance and is scaled with corresponding eigenvalues, i.e., $y_{i} = F x_{i}$, where $F$ is the top-left block of $(E\Lambda^{\frac{1}{2}})^{T}$ with dimension $D \times d$. Data reconstruction is performed via the operation of $F^{-1}$ on the obtained coefficients. 
\item
{\bf Independent Component Analysis (ICA)}: ICA attempts to decompose multivariate data into maximally independent non-Gaussian components.
Such a representation seems to be able to capture the essential structure of the data and provide a suitable representation which can be taken advantage of in neural networks \cite{hyvarinen2000independent}. FastICA, used here, introduced a different measure for maximizing the non-Gaussianity of rotated components \cite{hyvarinen1999fast}.

\item {\bf Tucker Decomposition}: The Tucker decomposition decomposes a tensor $T$ of order $N$ into a core tensor with the same order and $N$ unitary matrices. It is viewed as a higher order singular value decomposition (HOSVD). If we consider an image dataset as a tensor $T \in \mathbb{R}^{d_{1} \times d_{2} \times d_{3}}$ of order $3$, its Tucker decomposition is 
$T = \mathcal{T} \times_{1} U^{(1)} \times_{2} U^{(2)} \times_{3} U^{(3)}$, where $\mathcal{T} \in \mathbb{R}^{d_{1} \times d_{2} \times d_{3}}$ is the core tensor which, as a lower rank approximation of $T$, gives a representation basis for it. Unitary or factor matrices are two-dimensional matrices which help in  projecting $T$ into bases $\mathcal{T}$. The Tucker decomposition is widely used in compression by considering a subset of coefficients which carry most of the information in the dataset and eliminating lower rank coefficients, which typically has no adverse affect on tensor reconstruction.
\end{enumerate}

We also utilize the Discrete Wavelet Transform (DWT) as a representation basis for a ``held out" model. We use the DWT to set up a probabilistic model which can act as a basis for comparison for all generative approaches. The DWT offers a suitable and general basis for image representation which captures both frequency and location information. Therefore, it is highly viable as a benchmark model. We calculated DWT coefficients of datasets with a symmetric and biorthogonal $1.3$ scaling function. We trained a Gaussian mixture model on all DWT coefficients (DWT-GMM) except the block of high frequency coefficients. The DWT-GMM is used as a benchmark for all generative models and is not used as a separate generative approach (but we plan to explore this possibility in future work).

%% file: approach.tex
\vspace{-0.2 cm}
\section{Methodology}
\vspace{-0.3 cm}
\label{sec:methodology}
Our two stage approach comprising representation basis projection and deep learning is applicable to general 2D datasets. Below, we set up a cross-product of approaches wherein representation bases are paired with deep generative models. While we have elected not to explore variational autoencoders (VAEs) in the present work, this can be easily accommodated in the future.
\begin{enumerate}
\item
Data Projection:
We begin by projecting images and two-dimensional datasets into a representation basis space introduced in the previous section. Depending on the nature of the dataset, generative model, or representation basis approach, data preprocessing steps and some model customization are required to improve the results. In section~\ref{sec:experiments}, we explain the preprocessing steps or model specifications adopted for the datasets used in this study.
\item
Generative Modeling:
Generative models are applied to learn the distribution of a \emph{subset} of transformed coefficients obtained via one of the dimensionality reduction procedures detailed in section~\ref{sec:representation bases}. This is an efficient use of the compression aspect of the representation basis which makes the generative process fast and efficient. This way, the focus of generative modeling shifts from learning the distribution of data in pixel space to that of the distribution of coefficients in a more informative and structured space.
\end{enumerate}
The generative models used in this work for structured image generation are GANs, NFs, and GMMs. The reason for this choice of models is the different approaches they take towards learning the data distribution. These models are briefly outlined below. For detailed explanation of generative models and their variants, please see \cite{goodfellow2014generative,kobyzev2020normalizing}.
\vspace{-0.4 cm}
\subsubsection{GANs} 
are deep generative models which have shown promising results in generating high-resolution images \cite{goodfellow2014generative}. GANs are composed of two building blocks:  generator ($G$) and discriminator ($D$) networks which are trained in an adversarial fashion to defeat each other. A GAN is formulated as a minimax zero-sum game in which the generator and discriminator try to optimize the
value function $V$ from their own perspective:
\setlength{\belowdisplayskip}{4 pt} \setlength{\belowdisplayshortskip}{4pt}
\setlength{\abovedisplayskip}{4 pt} \setlength{\abovedisplayshortskip}{4pt}
\begin{equation}
\begin{split}
\min_{G} \max_{D} V(G, D) &= E_{y \sim p_{\mathbf{y}}} [\log D(y)]
+ E_{z \sim p_{\mathbf{z}}} [\log (1 - D(G(z)))], 
\end{split}
\end{equation}
where $p_{\mathbf{z}}$ is a predefined prior for the input noise variable $z$, and $p_{\mathbf{y}}$ is the true distribution of the data. Despite their impressive results on learning complex data distributions and generating natural-looking images, GANs cannot perform inference and evaluation of the probability density of new images and datasets---especially important in the domain of scientific datasets.
\vspace{-0.5 cm}
\subsubsection{NFs} were proposed as a generative model based on random variable transformations to approximate a tractable probability distribution such that sampling and inference is exact and efficient \cite{rezende2015variational}. The basic idea of NFs is to transform a simple probability distribution (typically Gaussian) into a complex one via learning a sequence of invertible and differentiable mappings (bijectors). This is the generative direction.
Applying a chain of mappings (bijectors) $f_{k}$, $k = 1, 2,\ldots, K$ on the random variable $\mathbf{z}_{0} \sim p_{0}(z_{0})$ results in a random variable $\mathbf{z}_{K} = f_{K} \circ f_{K-1} \circ \ldots \circ f_{1} (\mathbf{z}_{0})$ with probability distribution $p_{K}$:
\begin{equation}
p_{K} (\mathbf{z}_{K}) =  p_{0} (\mathbf{z}_{0}) \prod _{k = 0} ^{K}  \bigg | \text{det}\frac{\partial f_{k}}{\partial \mathbf{z}_{k-1}} \bigg |^{-1}.
\end{equation}

In order for these transformations to be practical, determinants of their Jacobians should be easy to compute. Some of the suggested approaches are
RealNVP \cite{dinh2016density}, Glow \cite{kingma2018glow}, and FFJORD \cite{grathwohl2018ffjord}. To implement NFs, we used the probability library of TensorFlow \cite{abadi2016tensorflow} and its distributions module. Bijectors were also trained by the FFJORD module in TensorFlow.
\vspace{-0.6 cm}
\subsubsection{The GMM} is a parametric  method for probability density function estimation. The density function is represented as a weighted sum of Gaussian components  \cite{reynolds2009gaussian}. The Gaussian mixture model represents data as normally distributed subpopulations with a hidden, unknown digital membership. The density of $X$ is formulated as a weighted sum of $K$ Gaussian distribution $N(\mu_{k}, \Sigma_{k})$ as follows:
\begin{equation}
   p(x|\pi,\mu,\Sigma) = \sum_{k=1}^{K} \pi_{k}N(x;\mu_{k},\Sigma_{k}); \;\; \text{with} \sum_{k=1}^{K} \pi_{k} = 1.
\end{equation}
The parameters of the GMM model are estimated by maximum likelihood estimation (MLE). Typically, an iterative Expectation-Maximization (EM) algorithm \cite{dempster1977maximum} is applied which turns out to be reasonably efficient for this MLE problem.

%% file: experiments.tex
\vspace{-0.3 cm}
\section{Experimental Results}
\vspace{-0.3 cm}
\label{sec:experiments}
In this section, we detail the experimental evaluation of the two-step process described in the previous section on two different datasets, one from image processing and the other from scientific computing. To compare the performance of generative models in representation bases, we executed a set of experiments on the cross product of models and representation bases for these datasets. 

With most image datasets, the pixel intensities range from $0$ to $255$ and are frequently normalized to a different range like $[0,1]$ for training. Unlike image datasets, scientific datasets are  not visually meaningful to human perception. Hence, measures like FID \cite{heusel2017gans} developed for image quality of GANs based on the Inception v3 model are not immediately applicable to scientific datasets like XGC, where each two-dimensional slice has a different range, so \emph{there is no unified range} in this dataset like there is in typical image datasets. Based on this observation, we propose a likelihood-based metric that we believe will be more suitable for this and other similar scientific datasets.

\vspace{-0.6 cm}
\subsubsection{Datasets.}
\begin{figure}[t]
\centering
\subfigure{
        \includegraphics[width=0.3\textwidth]{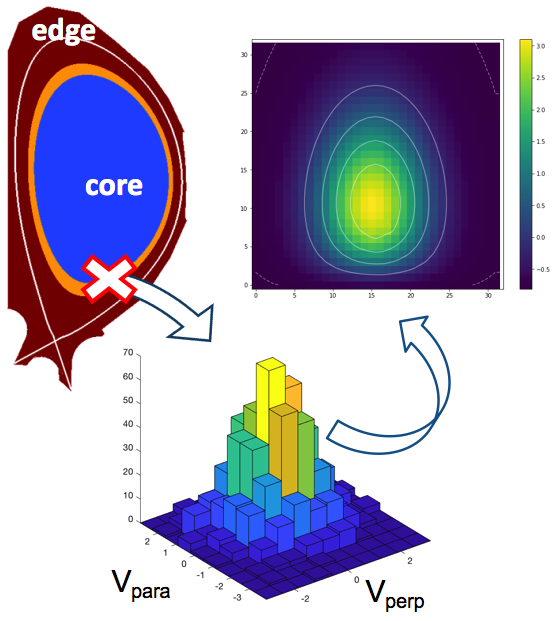}
    }
    \hspace{1 cm}
    \subfigure{
        \includegraphics[width=0.35\textwidth]{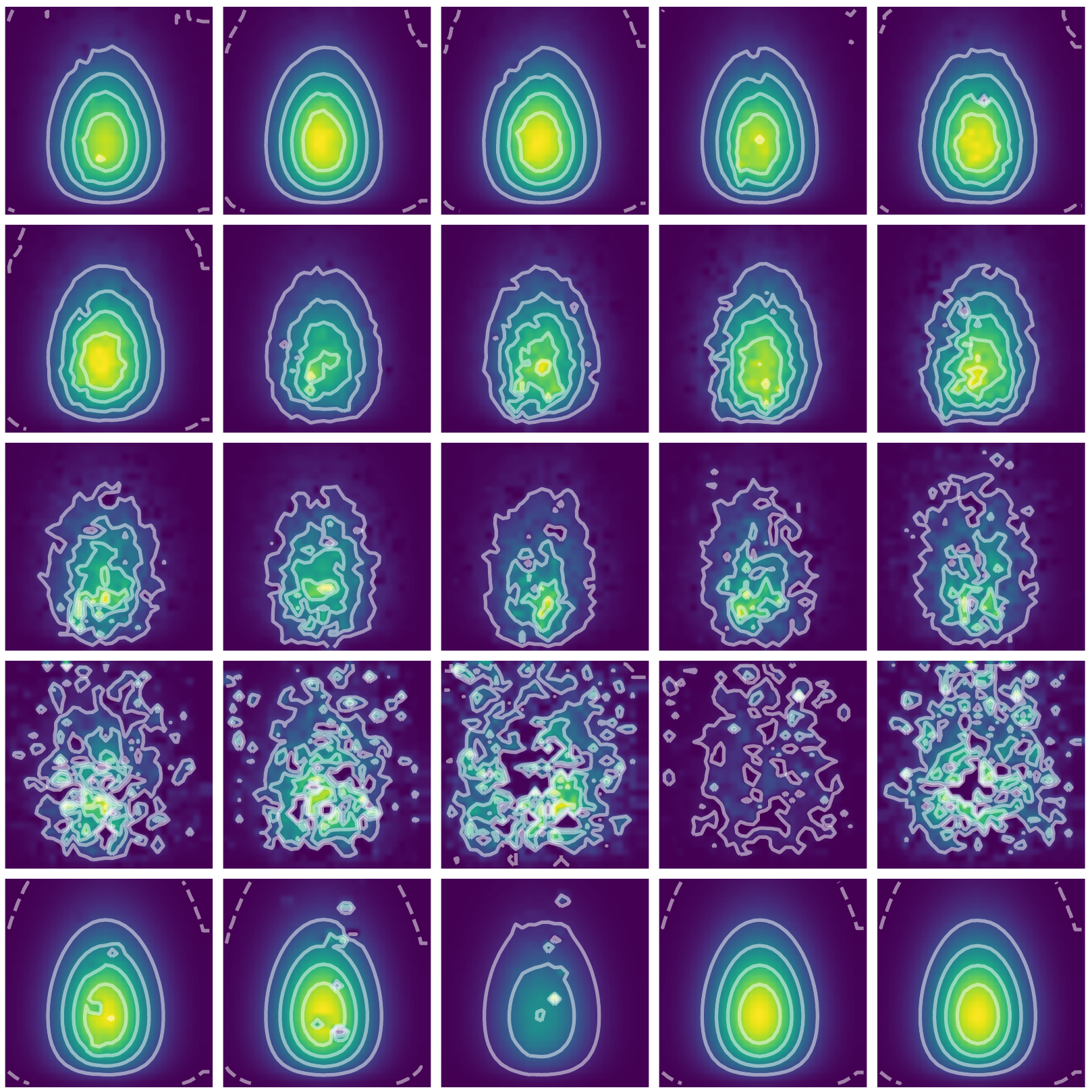}}
\caption{Depiction of a  node in the XGC dataset (left). The $x$ and $y$ axes of the histogram represent the perpendicular and parallel velocities of photons, respectively. Samples from dataset (right).}
\label{fig:XGC demo}
\end{figure}

We experimented with two datasets: Fashion MNIST \cite{xiao2017fashion} and XGC \cite{cole2019verification}. Fashion MNIST is a standard, curated and widely used dataset consisting of ten classes of clothing items. XGC consists of 16 planes corresponding to a doughnut's cross-sections. Each plane consists of $12,458$ nodes with each node representing a histogram of perpendicular and parallel velocities of photons at specific checkpoints (please see 
Figure~\ref{fig:XGC demo}). The histograms are not necessarily normalized. The velocity histogram of one of these nodes is depicted on the left in Figure~\ref{fig:XGC demo}. The goal is to derive a generative model to simulate the two-dimensional velocity histograms of particles which are represented as images in a compressed and efficient way.

\begin{figure*}[t]
\centering
    \subfigure{
        \includegraphics[width=0.42\textwidth]{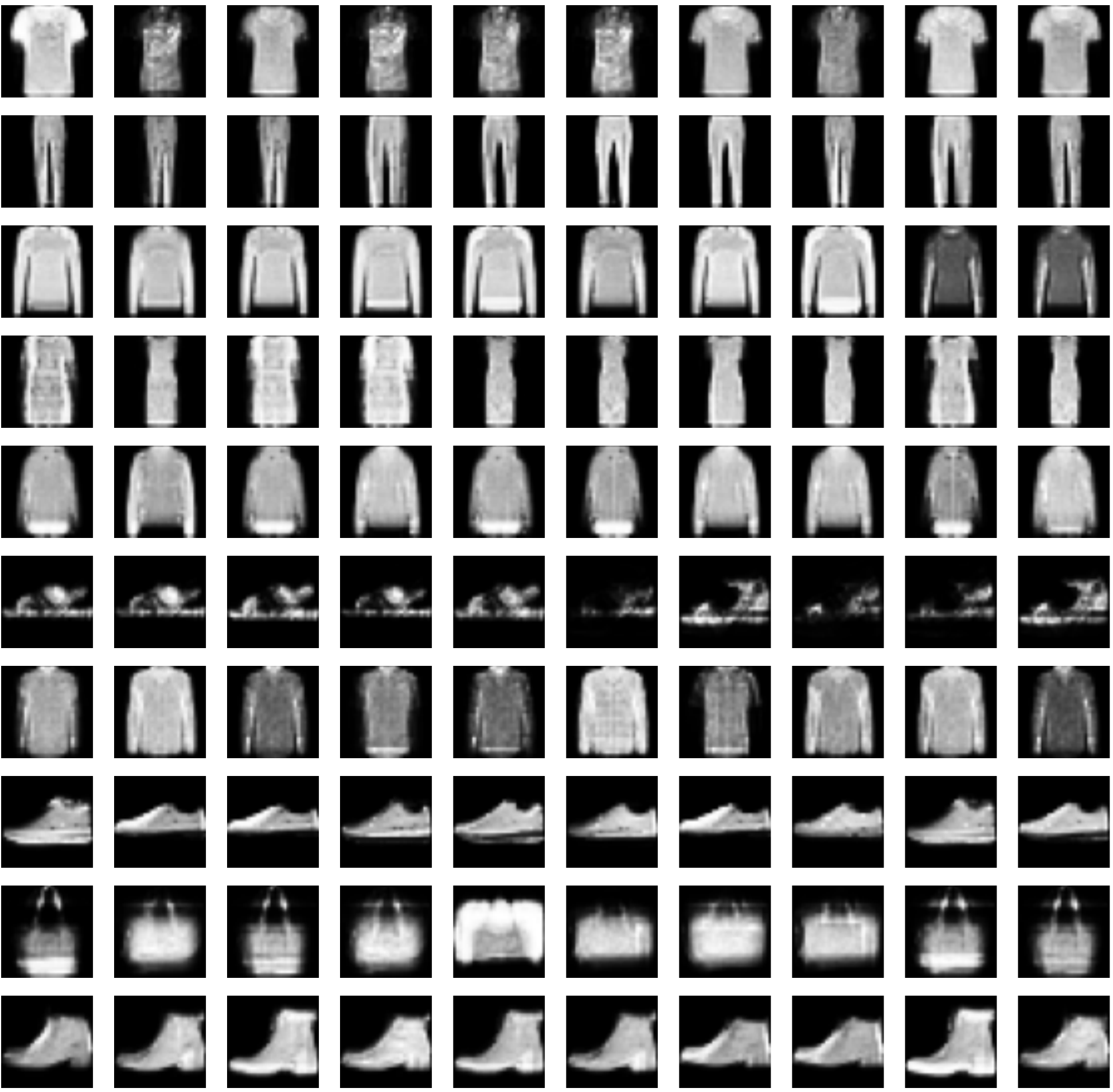}
    }
    \vspace{-0.01 cm}
    \subfigure{
        \includegraphics[width=0.42\textwidth]{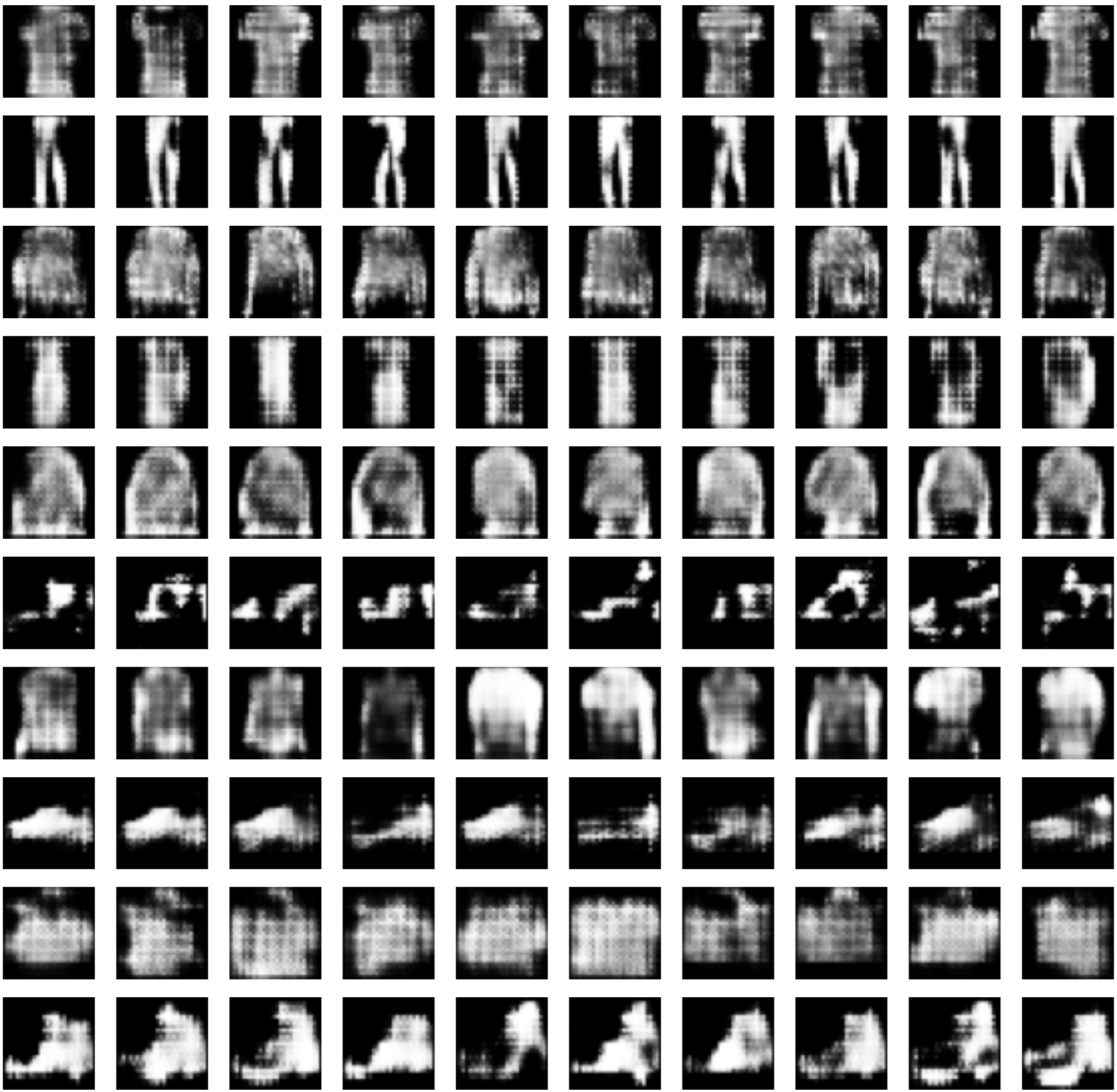}} 
        \vspace{-0.01 cm}
\subfigure{
        \includegraphics[width=0.42\textwidth]{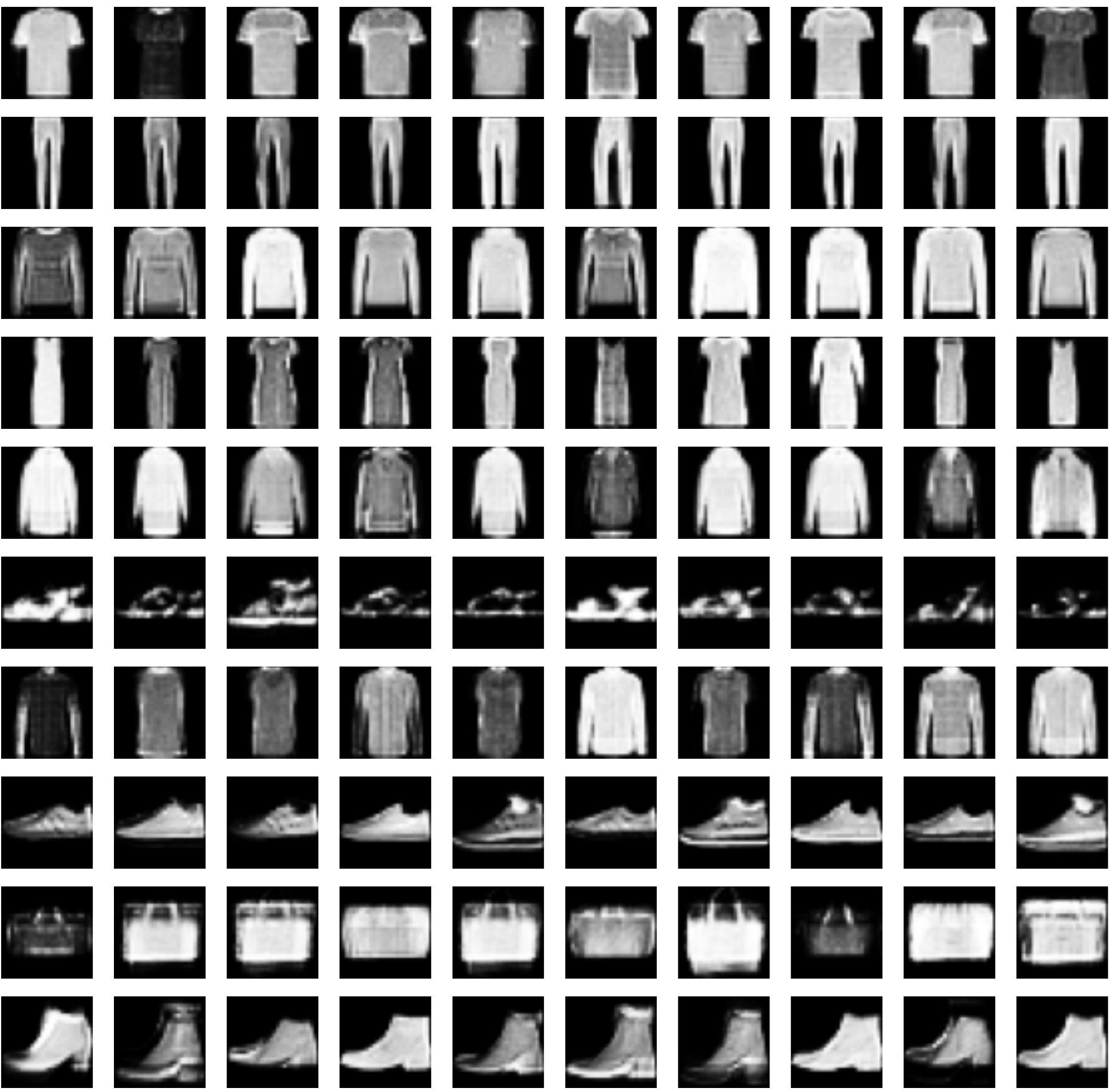}}
    \hspace{0.001 cm}
\subfigure{
        \includegraphics[width=0.421\textwidth]{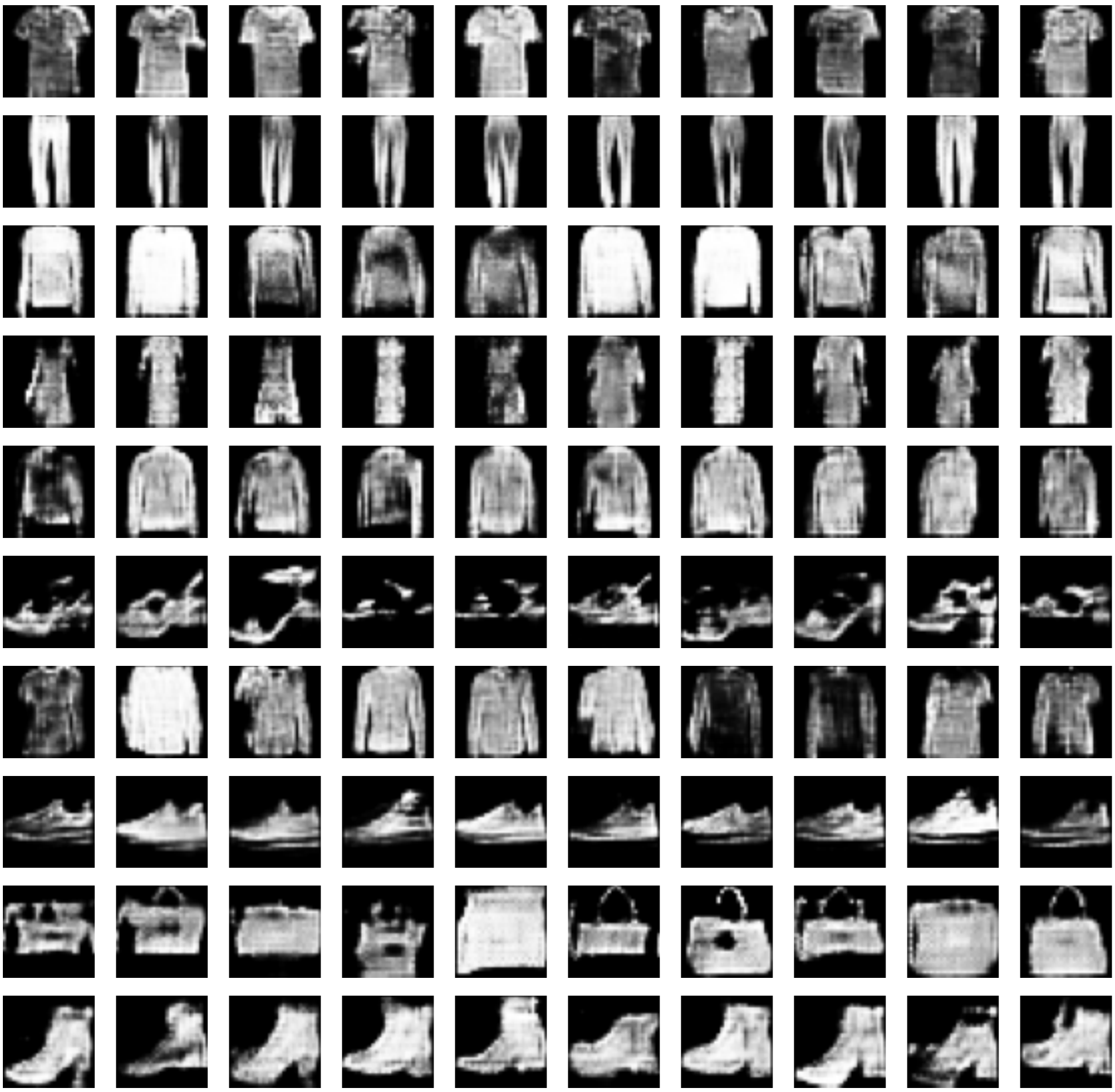}}
        \caption{Fashion generated images via ICA-GAN (left) and pixel-GAN (right) at epoch 10 (upper row) and epoch 50 (lower row). Image samples are randomly drawn and not cherry-picked. ICA-GAN generates plausible images close to the dataset from early iterations, with much fewer artifacts in terms of shape and texture. Pixel-GAN takes many more iterations to converge, with some images having artifacts.}
\label{fig:fashion-10-50}
\end{figure*}
\vspace{-0.6 cm}
\subsubsection{Preprocessing.}
Slightly differing approaches were taken for the two datasets for projection to the PCA/ICA basis. Since the ICA bases generate an unconstrained range of pixel values, for the Fashion dataset, we first project the image intensities from a discrete range of $[0, 255]$ to the continuous range $[0, 1]$ and then project these intensity values to a wider range using an inverse sigmoid function $ y(x) = \log \frac{x}{\beta (1-x)}$, where $\beta$ is a gradient slope factor. PCA/ICA is then applied to this new range of values. Since the inverse sigmoid is not defined at $0$ and $1$, we map all intensities to the interval $[\epsilon, 1-\epsilon]$ for some value of $\epsilon$ and then apply the inverse sigmoid before applying PCA and ICA. In our experiments, $\epsilon$ is set to $0.001$. The inverse sigmoid is used because the last nonlinear activation function in the generator of the GAN architecture for this dataset is sigmoidal, and therefore, we can match the intensities to the training data. 

For the XGC dataset, the values are normalized numbers of particles in simulation which have a specific perpendicular and parallel velocity at a checkpoint. These values are normalized by the mean and standard deviation of each image separately (essentially a per image $Z$ score). 
Because each image has a different range, it impacts the choice of architecture and activation function in generative models. 
\vspace{-0.6 cm}
\subsubsection{Generative Models Architecture.} 
\begin{wrapfigure}{}{0.6\textwidth}
\centering
\includegraphics[width=0.45\textwidth]{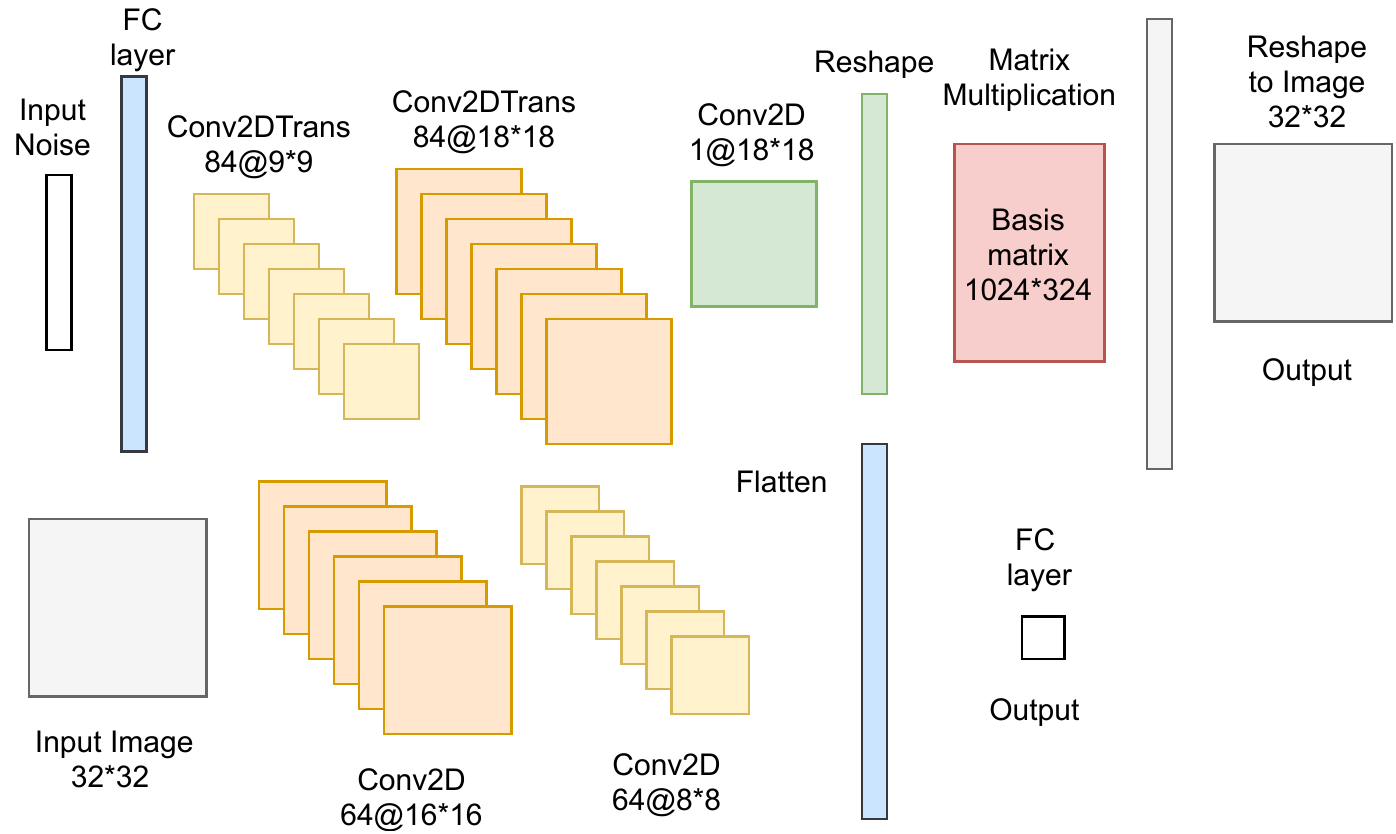}
\caption{The GAN architecture for the XGC dataset; generator upper row and discriminator lower row. A subset of representation bases is chosen.}
\label{fig:gan-arch}
\end{wrapfigure}

Figure~\ref{fig:gan-arch} depicts the architecture of a GAN for the XGC dataset. We used upsampling (conv2DTranspose) layers in the generator with a linear activation function in the last layer to allow for the range of generated images to be chosen freely for each image. This way, the generator is constrained to learn coefficients such that,  after image reconstruction (via the transformation matrix), the values follow an acceptable range that is similar to the training set. The GAN architecture for Fashion is very similar to XGC, except for the number of filters and the use of sigmoidal activation instead of linear activation at the last layer of the generator.

The advantage of representation bases and dimensionality reduction is more tangible in NFs because these models are computationally expensive: when input/output dimensions are increased, the number of training parameters grows rapidly. We considered four layer of bijectors and 50 additional nodes in the hidden layers of bijectors. 
\vspace{-1.01 cm}
\subsubsection{Dimensionality Reduction.}
 Our goal is to learn the distribution of a subset of coefficients as an efficient approach to data generation. The number of top eigenvalues and corresponding eigenvectors for each dataset was determined based on the $\ell_2$ distance between the training dataset and reconstructed images. For Fashion MNIST, more coefficients were needed to meet a certain error threshold: $324$ and $400$ coefficients were chosen respectively for XGC and Fashion datasets. We compare the performance of generative models in learning the distribution of a subset of coefficients for each method via different measures. We observed that for XGC, PCA, and ICA have similar performance across the board. For Fashion however, ICA had better performance than PCA, and therefore, we only focused on the performance of ICA-GAN for this dataset.
\vspace{-0.55 cm}
\subsubsection{Metrics.}
Many qualitative and quantitative measures have been proposed to evaluate generated images and learned probability distributions of generative models \cite{borji2019pros}.
\vspace{-0.05 cm}
We consider two conventional and widely-used quantitative metrics---Frechet inception distance (FID) and average log-likelihood (entropy) of samples in kernel density estimation (KDE) \cite{goodfellow2014generative} for evaluating the learned distributions and generated samples from different models. 
FID was proposed as a statistical metric to measure the similarity between two distributions. First, by running the Inception v3 net on real and generated images, high level features (pool3 layer) are extracted as an embedding for images, and then a separate multivariate Gaussian distribution is fitted to real and generated embeddings. 
FID does not seem to be a suitable metric for evaluating scientific dataset generation. The features extracted from a deep learning network trained on real images which
are perceptually meaningful to human vision \emph{are not necessarily appropriate} for scientific data (where perceptual quality is not used). 
Furthermore, FID only provides a single scalar measure for the entire dataset and does not take the actual likelihood of the training set or generated images into account. For these reasons, we resort to metrics based on probability distributions and the likelihood of generated samples with respect to a reference model. KDE directly fits a probability density model to the generated images. We calculated the average of the negative log-likelihood (NLL) i.e. entropy of images sampled from learned distributions with respect to this reference density model (Table~\ref{tab:generative_models_results}). However, as mentioned above, it is much more efficient to learn probability distributions in the space of a representation basis than in pixel space. These considerations affect our choice as described below.
As mentioned in section~\ref{sec:representation bases}, we consider a reference model which is essentially a GMM on DWT coefficients. This model serves as a benchmark and is \emph{not used} for generation or sampling. The number of coefficients used is $3 \times 256$ and $3 \times 324$ coefficients for Fashion and XGC datasets, respectively. To assess the learned density distributions via different models, we use the average of NLL values of sampled images from learned distributions via GANs, GMM, and NFs in the DWT-GMM space---essentially the DWT entropy. Furthermore, we compute the $\ell_1$ distance between the density curves obtained via KDE of the NLL values of generated images in the DWT-GMM model (see Figures~\ref{fig:Fashion DWT NLL} and \ref{fig:XGC DWT NLL}). Essentially, this distance is computed between the density curve of each model and the density curve of the real dataset on the interval that contains most of its density volume.

%% file: discussion.tex
\vspace{-0.4 cm}
\subsubsection{Results and Discussion}
\label{sec:discussion}
\begin{table*}[t]
\centering
\caption{Image generative models using representation bases with dimensionality reduction (324 and 400 coefficients: PCA, ICA, Tucker for XGC and Fashion datasets respectively). Numbers $10$ and $50$ in the $4^{\text{th}}$ column denote the learned distribution at that epoch number. Metrics: DWT and KDE entropies (DWT-E and KDE-E respectively), FID and the $\ell_1$ distance (scaled by 10$^{-2}$). For all metrics, lower is better.}
{\normalsize
\begin{tabular}{|c|c|c|c|r|r|r|r|}
\hline
Dataset                 & Model                 & Loss                 & Target Dist. & DWT-E & KDE-E  & FID & $\ell_1 $ \\ \hline
\multirow{8}{*}{XGC}     & \multirow{4}{*}{GAN} & \multirow{4}{*}{ADV} & ICA                 &         $-2,239$    &   $-82$  &   $5.6$  & $4.0$ \\ \cline{4-8} 
                         &                       &                 & PCA                 &        $-2,261$     &    $-76$ &  $5.6$   & $4.0$  \\ \cline{4-8}
                       &                      &                      & Tucker              &     $25,745$   &  $-33$   &   $13.9$  & $5.0$ \\ \cline{4-8} 
                         &                       &                      & Pixel               &        $311,895$    &  $207$   &  $2.8$    &  $5.0$ \\ \cline{2-8} 
                         & \multirow{2}{*}{NF}  & \multirow{2}{*}{MLE} & ICA                 &       $3,241$      &   $477$  &  $6.0$   &  $4.8$\\ \cline{4-8} 
                         
                         &                   &                      & Tucker              &      $1,682$         &  $1,105$    &  $26$    & - \\ \cline{2-8} 
                         & \multirow{2}{*}{GMM}                   &\multirow{2}{*} {MLE}                 & ICA                 &        $-1,096$     &  $39$  &   $5.5$  &  $4.3$\\ \cline{4-8} 
                         &          &       &Tucker        &         $-2,474$   &       $-455$      & $2.4$ &  $3.0$\\
                         \hline \hline
\multirow{8}{*}{Fashion} & \multirow{5}{*}{GAN} & \multirow{5}{*}{ADV} & ICA 10                 &      $-2,451$       &  $-453$   &  $4.9$  & $2.7$ \\ \cline{4-8} 
                         &                       &          & ICA 50                 &    $-2,550$         &    $-450$ &   $2.2$   &  $2.2$ \\ \cline{4-8} 
                          &                      &                     & Pixel 10            &     $-1,576$     &  $-307$   &    $5.0$ & $3.5$ \\\cline{4-8} 
                         &                       &                      & Pixel 50            &    $-1,841$     &    $-346$ &  $1.3$ & $2.6$ \\ \cline{4-8}
                         &          &    &   Tucker   &   $-1,146$   &  $-333$   & $2.2$ & $4.4$
                         \\\cline{2-8} 
                         & \multirow{1}{*}{NF}  & \multirow{1}{*}{MLE} & ICA                 &    $-1,078$         &  $-329$   &   $2.9$ & $8.9$\\ \cline{2-8} 
                         
                         & \multirow{2}{*}{GMM}  & \multirow{2}{*}{MLE}                  & ICA                  & $-1,033$                &     $-361$      &  $2.3$ & $5.7$ \\\cline{4-8} 
                         &      &   & Tucker &  $-2,660$ &  $-480$  & $21$ & $2.6$ \\\hline
\end{tabular}
}
\label{tab:generative_models_results}
\end{table*}
Experimental results for different combinations of modeling and generative bases are provided in Table~\ref{tab:generative_models_results} for the two datasets. These results show that ICA-GANs preserve the statistical properties of each dataset despite higher FID scores compared to equivalent pixel-GANs. The \emph{very high entropy} of pixel-GAN for the XGC dataset shows that the learned distribution of data via pixel-GAN is far from the true distribution despite generating images which are visually similar to the training set. This indicates that generating a scientific dataset in pixel space may not be a reasonable approach. ICA-GAN (at epoch 50) had the best performance on the Fashion dataset which is a curated dataset with images being approximately registered within classes. Note that we cannot expect scientific data to be pre-registered. The better performance of the pixel-GAN on Fashion compared to XGC is partly because of the unified range of pixel intensities for Fashion allowing for the use of a single sigmoid activation function which confines the generated pixel values within $[0, 1]$. Overall, the results also show that GMMs with a representation basis (after preprocessing) are powerful  generative models for two dimensional datasets, regardless of whether the data arise from standard imagery or from scientific simulation. 

\begin{figure*}[t] 
    \centering
    \includegraphics[width=\textwidth]{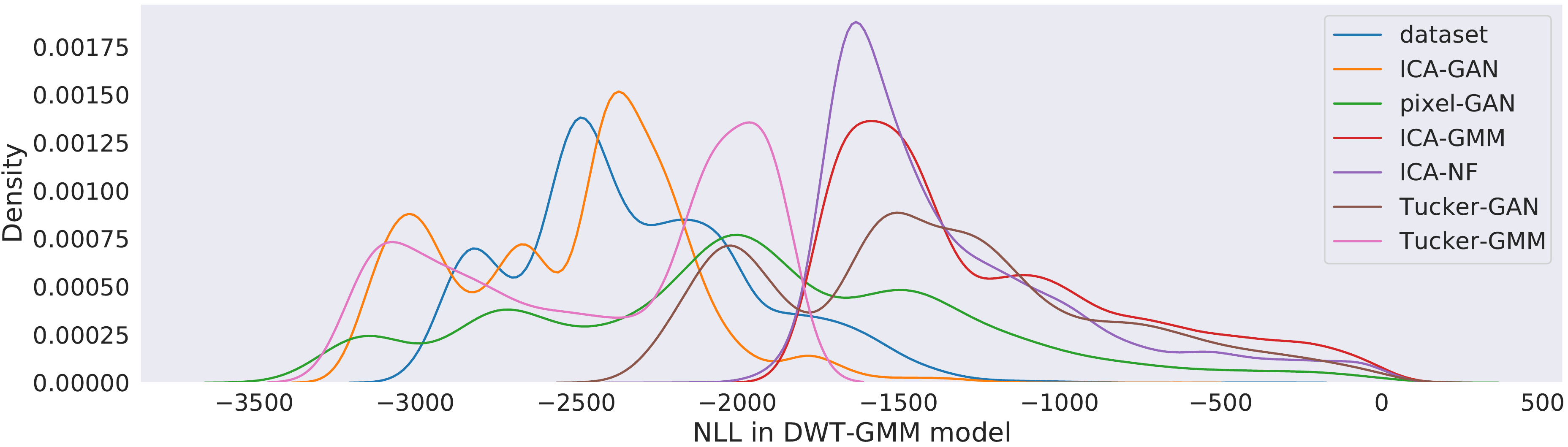}
    \caption{Fashion negative log-likelihood (NLL) density distributions in the DWT-GMM benchmark model. For better demonstration, only samples with negative NLL are plotted (data density concentration). ICA-GAN samples depict NLL values near NLL of Fashion dataset.}
    \label{fig:Fashion DWT NLL}
\end{figure*}

\begin{figure*}[t] 
    \centering
    \includegraphics[width=\textwidth]{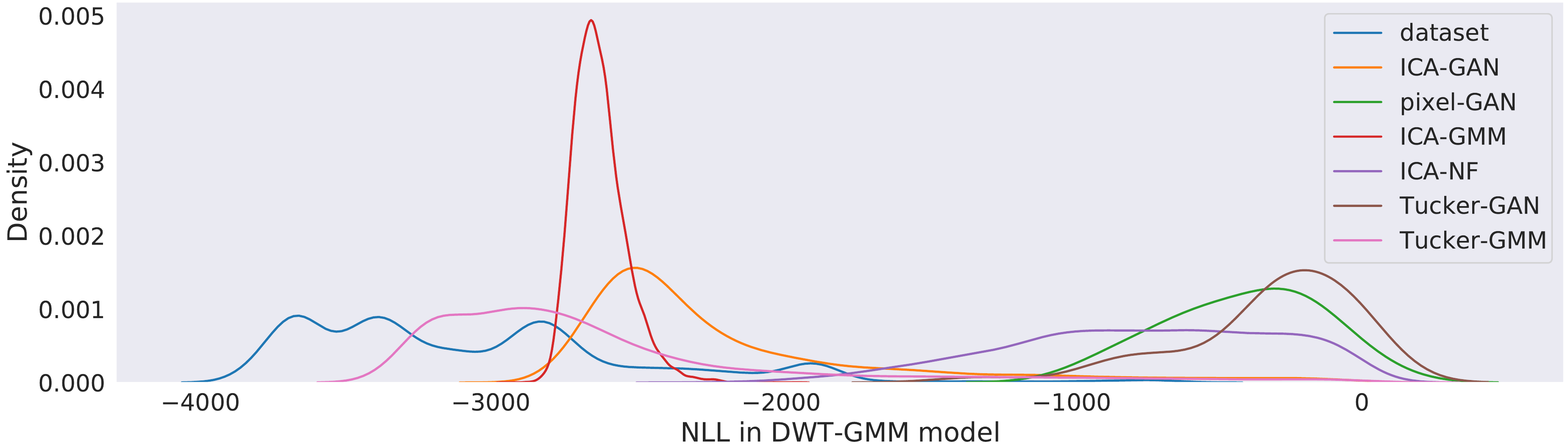}
    \caption{XGC negative log-likelihood (NLL) density distributions in the DWT-GMM benchmark model. ICA-GMM seems to have better NLLs (lower entropy) however, GANs perform better on average since ICA-GMM has only part of samples ($7\text{K}$ out of $12\text{K}$) in negative range.}
    \label{fig:XGC DWT NLL}
\end{figure*}

Figure~\ref{fig:fashion-10-50} shows that with a reasonable and simple architecture of GANs for learning the distribution of ICA coefficients of the Fashion dataset, it is possible to generate plausible looking images which are mostly texture and shape artifact-free from early iterations. Furthermore, Figure~\ref{fig:Fashion DWT NLL} (and the $\ell_1$ distance in Table~\ref{tab:generative_models_results}) also indicate that the generated images by ICA-GAN has the closest entropy to the Fashion dataset in both DWT-GMM and KDE benchmark models among deep generative models and representation bases. The pixel-GAN on the other hand does not produce close-to-dataset images until later iterations (with an identical discriminator) while many of the images have artifacts in terms of shape and texture. From Figure~\ref{fig:XGC DWT NLL}, it might seem that the GMM has a better performance compared to the ICA-GAN for XGC data. However, it is important to note that less than $\frac{2}{3}$ of the ICA-GMM samples fall in the negative range of NLL while ICA-GAN shows a more homogeneous behaviour and hence lower entropy. 

%% file: conclusion.tex
\vspace{-0.8 cm}
\section{Conclusions}
\label{sec:conclusion}
\vspace{-0.2 cm}
We proposed a framework for fast and efficient image generation that combines a representation basis approach with deep generative modeling. Our rationale was that learning a basis for data which preserves the statistical structure and correlation among image pixels can be a useful preprocessing step for the development of generative models. Furthermore, representation bases can be deployed for data compression during generation which is a boon for computationally intensive generative modeling frameworks. Immediate future work will focus on using over-complete dictionaries and coefficient compression within generative modeling.
\vspace{-0.2 cm}
\section*{Acknowledgments} 
\vspace{-0.2 cm}
This material is based upon work supported by the U.S. Department of Energy, Office of Advanced Scientific Computing Research, Scientific Discovery through Advanced Computing (SciDAC) program under Award Number DE-SC0021320.